\documentclass[fleqn,10pt]{wlscirep}
\usepackage[utf8]{inputenc}
\usepackage[T1]{fontenc}
\usepackage{subcaption}
\usepackage{tikz}
\usepackage{lipsum}

\title{The Way Up: A Dataset for Hold Usage Detection in Sport Climbing}

\author[1,*]{Anna Maschek}
\author[1,+]{David C. Schedl}
\affil[1]{University of Applied Sciences Upper Austria, Campus Hagenberg, Austria}

\affil[*]{anna.maschek@fh-hagenberg.at}

\affil[+]{david.schedl@fh-hagenberg.at}

\begin{abstract}
Detecting an athlete’s position on a route and identifying hold usage are crucial in various climbing-related applications. 
However, no climbing dataset with detailed hold usage annotations exists to our knowledge. 
To address this issue, we introduce a dataset of 22 annotated climbing videos, providing ground-truth labels for hold locations, usage order, and time of use. 
Furthermore, we explore the application of keypoint-based 2D pose-estimation models for detecting hold usage in sport climbing. We determine usage by analyzing the key points of certain joints and the corresponding overlap with climbing holds.
We evaluate multiple state-of-the-art models and analyze their accuracy on our dataset, identifying and highlighting climbing-specific challenges. 
Our dataset and results highlight key challenges in climbing-specific pose estimation and establish a foundation for future research toward AI-assisted systems for sports climbing.
\end{abstract}
\begin{document}

\flushbottom
\maketitle

\thispagestyle{empty}

\section*{Introduction}
\label{sec:intro}

Sport climbing has seen a rise in popularity in recent years, with its inclusion in the 2020 Olympics further increasing its global visibility. 


Detecting an athlete's position on a route and identifying hold usage are crucial for automated competition scoring \cite{michenthaler_automated_2022}, performance and movement analysis \cite{boulanger_automatic_2015}, and other climbing research. However, these tasks are often performed manually or with dedicated hardware, limiting scalability and accessibility.
To address these challenges, we explore using computer vision algorithms to automatically detect a climber’s position on a route, explicitly identifying the holds in use. By integrating this information with the route's predefined hold positions, such a system could support automated feedback and training applications.

We employ keypoint-based 2D-pose-estimation models to detect joint locations and to determine hold usage. 
While pose estimation has been widely studied in traditional sports, climbing remains underexplored due to its unique challenges, such as frequent self-occlusions and non-standard movement patterns. To our knowledge, no prior work has systematically evaluated multiple pose estimation models for detecting hold usage in climbing.
To objectively compare different pose-estimation models, we created a dataset of 22 annotated climbing videos, including hold locations, usage order, and time of use, to facilitate future research in climbing-specific pose estimation and movement analysis.

To enable future development of real-time, portable climbing assistance tools - particularly for visually impaired climbers - our work focuses on approaches that are compatible with commonly available hardware. This application-oriented perspective guided both the design of the dataset and the choice of models and processing steps.

The contributions of this paper are as follows:
\begin{itemize}
\item A publicly available dataset of 22 annotated climbing videos, including hold locations, usage order, and time of use \cite{ourdata}. 
\item An evaluation of state-of-the-art 2D-pose-estimation models for detecting hold usage in sport climbing.
\end{itemize}

\section*{Related work}
\label{sec:relatedwork}

With the increasing popularity of climbing, there has been a surge in research focusing on the sport.

\subsection*{Motion detection in climbing}
In a review by Andric et al. \cite{andric_sensor-based_2022}, 43 studies related to climbing were analyzed. This review identified three distinct approaches for detecting climber motion: embedded sensors, body-worn sensors, and external sensors.

Embedded three-axis force sensors, have been used by Pandurevic et al. \cite{pandurevic_methods_2019} in climbing holds to detect force peaks on hands and feet while climbing. They combine this with a depth camera-based motion tracking system to provide information about the climber's body positioning.

Body-worn sensors, such as inertial measurement units (IMUs), have been used in various applications. For example, Boulanger et al. \cite{boulanger_automatic_2015} detected four different climbing states — immobility, hold exploration, pelvis movement, and global motion — while Kosmalla et al. \cite{kosmalla_climbsense_2015} achieved automated recognition of climbing routes. 
Moreover, Seifert et al. \cite{seifert_role_2017} combined eye-tracking glasses with IMUs to assess the impact of route previewing strategies on climbing fluency and limb movement.

Other works have explored the use of external sensors, such as camera-based systems, for analyzing climbing movements.

One area of interest is the relationship between a climber's gaze and movement \cite{2024_Vrzáková_ObjectDetectionActivityRecognition, Pantry2024DeepLF}. Vrzáková et al. \cite{2024_Vrzáková_ObjectDetectionActivityRecognition} used a custom trained YOLOv5 model to determine holds and grasped holds from a video recorded from the climber's head position. This is combined with information from gaze tracking hardware to determine the relation between the climber's gaze and body movement.

Beltrán et al. \cite{beltran_beltran_climbing_2023} proposed a computer vision-based approach to automatically detect six common movement errors in novice climbers. Their method employs Apple's Vision framework to extract 2D skeletal keypoints, which are then combined with LiDAR depth data for 3D joint localization. By analyzing joint motion, they classify the climber's movement phase, enabling the identification of typical errors associated with each phase.

Speed climbing has been analyzed using computer vision approaches to compare different climbers\cite{pandurevic_analysis_2022, petr2021Speed21}. Pandurevic et al. \cite{pandurevic_analysis_2022} applied OpenPose to speed-climbing footage to extract climbers' keypoints. These keypoints were then used to analyze various movement characteristics, including the position, velocity, and acceleration of the center of gravity (COG), joint angles, and contact times. Additionally, the study measured the time taken to transition between specific holds. Petr et al.\cite{petr2021Speed21} used YOLOv3 for person detection in combination with HRNet for 2D human pose estimation to analyze climbers' movement styles. By leveraging a specialized variant of Dynamic Time Warping (DTW) that aligns movements across key body parts, different runs of the same climber could be matched.

Michenthaler et al. \cite{michenthaler_automated_2022} also employed YOLOv3 to detect climbers on the wall automatically. They used these detections to determine the holds used by these climbers based on the coverage of the holds, aiming to automate scoring in climbing competitions.

Ludford\cite{Ludford2024DevelopmentOA} presents the development of a gym-wide bouldering performance analysis tool that utilizes computer vision on footage from multiple cameras within the gym. The system employs YOLOv7 for pose estimation and a custom-trained YOLOv7 model for hold detection to determine the number of routes climbed and the number of attempts per route for each climber. 

Climb-o-Vision \cite{richardson_climb-o-vision_2022}, a system to guide visually impaired climbers, uses YOLOv5 to detect climbing holds on the wall. Subsequently, the positions of these holds are relayed to the climber using an electrotactile tongue interface.

Our work focuses on a camera-based approach, as it allows climbers and gyms to leverage existing hardware without requiring specialized or costly equipment. Pose estimation algorithms will be used to detect climbers' movements. Notably, there hasn't been a detailed evaluation of multiple pose estimation algorithms for sport climbing, prompting our investigation in this area.

\subsection*{Climbing datasets}
Some datasets have been introduced to support research at the intersection of climbing and computer vision, addressing tasks such as hold detection, pose estimation, and gaze tracking.
A publicly available dataset \cite{climbing-holds-and-volumes_dataset} provides labeled images of climbing holds and volumes, supporting hold segmentation and recognition tasks.
Yan et al. \cite{yan2023cimi4d} introduced a 3D climbing motion dataset focused on 3D pose estimation. While it includes scene contact annotations indicating which holds were involved, it lacks the temporal and order information of hold usage, limiting its applicability for research focused on hold interaction.
Vrzáková et al. \cite{2024_Vrzáková_ObjectDetectionActivityRecognition} describe a dataset for hold and grasp detection on footage recorded with head-mounted devices. The data was used to train a custom YOLOv5 model to associate the climber’s gaze with their grasping actions, enabling research on the relationship between visual focus and climbing movements.
Similarly Pantry et al. \cite{Pantry2024DeepLF} presented a dataset linking gaze points with climbing holds. This dataset allows for the study of how climbers visually engage with holds before and during movement.
The SPEED21 dataset \cite{petr2021Speed21} focuses on speed climbing, providing 2D skeleton data with 16-joint coordinates extracted from 362 speed climbing performances. The video sequences for the dataset were obtained from competition videos on the official IFSC YouTube channel. Annotations were obtained with YOLOv3 for person detection and pose estimations from HRNet.

These datasets contribute to different aspects of climbing research, from hold detection and pose estimation to motion analysis and gaze tracking. In contrast, our dataset provides temporal information, including the order in which holds are used during a climb. This temporal sequencing is essential for modeling climbing movements and has not been addressed in existing datasets.

\section*{Dataset}
\label{sec:dataset}

Accurately identifying the holds a climber is using at any given moment is crucial for climbing-related research \cite{michenthaler_automated_2022,boulanger_automatic_2015, 2024_Vrzáková_ObjectDetectionActivityRecognition}.
We propose an annotated dataset of 22 climbing videos showing 10 athletes. Our annotations specify the time intervals during which each hold is used and thus allow for the systematic comparison of different approaches. To the best of our knowledge, no existing dataset in the literature provides this level of annotation for climbing.

\subsection*{Data acquisition}
\begin{figure}
  \centering
  \begin{subfigure}{0.289\linewidth}
    \includegraphics[width=\linewidth]{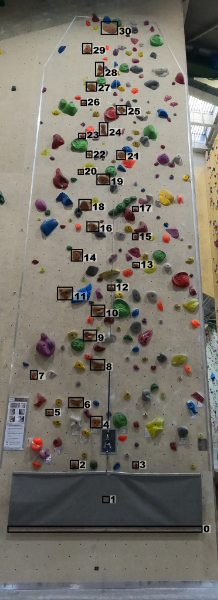}
    \caption{orange route}
    \label{fig:routes-orange}
  \end{subfigure}
  \begin{subfigure}{0.29\linewidth}
    \includegraphics[width=\linewidth]{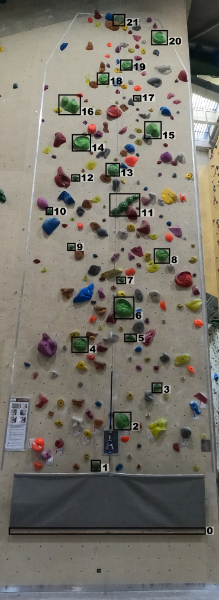}
    \caption{green route}
    \label{fig:routes-green}
  \end{subfigure}
  \caption{The two routes used in creating the dataset, annotated with the unique identifier assigned to each hold of the respective route.}
  \label{fig:routes}
\end{figure}

Participants were asked to climb two different routes on an 8.5-meter high, 0-degree inclined indoor wall, both of which were within the climbing abilities of the participants. The first route visualized in \autoref{fig:routes-orange}, referred to as the \textit{orange route}, had a difficulty rating of 4c. It featured jugs for all holds and short movements, with holds placed relatively close together. The second route visualized in \autoref{fig:routes-green}, referred to as the \textit{green route}, had a difficulty rating of 5a and included a mix of jugs and more challenging holds in the upper section, precisely two slopers or edges, depending on how they were used. This variation added a challenge for some climbers — the green route required longer, more dynamic movements due to the greater distance between holds.

\begin{figure}[t]
  \centering
   \includegraphics[width=0.5\linewidth]{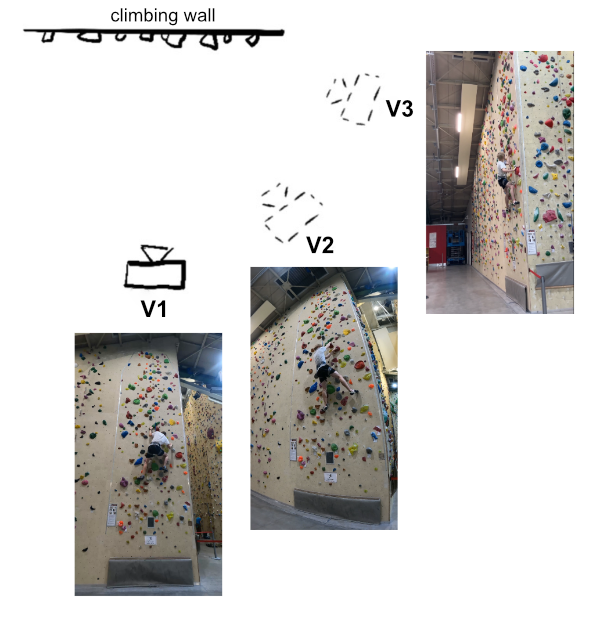}
   \caption{Different camera positions (v1, v2, v3) evaluated for dataset collection, with v1 selected as the final viewpoint.}
   \label{fig:cameraAngles}
\end{figure}

Filming was conducted using a handheld GoPro Hero 9 camera positioned at a 90-degree angle to the wall. This angle was chosen after preliminary tests, shown in \autoref{fig:cameraAngles}, to capture the climber's movements best. 
The videos were recorded in portrait mode, allowing the entire route to fit into the frame at all times, with a resolution of 3840x2160 pixels and 50 FPS. 

Additionally, a two-camera setup was tested, with one camera filming from behind and another mounted on the climber’s chest to capture hand positions obscured by the climber’s upper body. However, the chest-mounted camera often became tangled in the auto belay rope, which limited the climber’s movement and posed a risk of the camera falling. As a result, a single, perpendicular camera setup was selected for all subsequent recordings to ensure a natural climbing flow and safety for the climbers.

\subsection*{Participants information}
Ten climbers participated in the dataset collection, ensuring diversity in height (156–183 cm, mean 173.5 cm) and climbing experience. Ages ranged from 18 to 54 years, with an equal distribution of five females and five males. Climbing experience varied from a first-time climber to over fifteen years of regular training, with most participants primarily training indoors and favoring bouldering.
No clothing requirements were enforced; however, one participant was recorded twice, once wearing black clothing to evaluate potential issues with dark clothing. Participants were allowed to inspect the routes beforehand. Some had prior experience with the routes or had watched previous ascents, while others climbed on-sight.

Recordings took place over four days. On one occasion, the gym was busy, resulting in background movement in some videos. Additionally, a technical issue prevented recording the first two moves of one climb.
All participants provided informed consent for filming and dataset publication.

\subsection*{Data pre-processing}
After recording, all videos were downsampled to 25 FPS and resized to 720×1280 pixels to improve processing efficiency. Non-relevant segments were removed, including moments when climbers were preparing to start or being lowered after completing the climb. The audio track was also discarded as it contained no relevant information.
To enhance video stability, DaVinci Resolve's\footnote{\url{https://www.blackmagicdesign.com/de/products/davinciresolve} version 18.6} video stabilization feature was applied using perspective mode with the camera lock option enabled.
Finally, the faces of non-participating individuals appearing in the video were blurred to protect privacy.
The final dataset contains the original 4K videos at 50 FPS. While the provided annotations correspond to the downsampled videos, they can be transformed to align with the original-resolution videos.

\subsection*{Data annotation}
Video annotation was performed by an experienced climbing coach with a background in competition climbing, ensuring accurate labeling of hold usage and movement patterns. Each hold in both routes was annotated and assigned a unique identifier, as shown in \autoref{fig:routes}. 

Based on the hold identifier, hold usage was annotated for each climber in each video, in the following format. 
Each data entry specifies the climbers' extremity used to grab the hold and the duration (start and end in frames) the climber held onto a particular hold. Furthermore, extremity occlusions are recorded.
The extremity abbreviations are as follows: The first letter denotes the side (l for left, r for right), while the second letter indicates the extremity (h for hand, f for foot). Thus, this leads to 4 possible variants (lh, lf, rh, rf). The hold corresponds to the hold identifier from the route topo. If the climber used the wall instead of a designated hold, a \textit{w} was recorded as a placeholder for the hold number. For instance, extremity rh and hold 4 indicates that hold four was used by the right hand (as shown in \autoref{tab:annotatedFramesFormat}).

\begin{table}
\centering
\begin{tabular}{lllll}
\toprule
\textbf{ext.} & \textbf{hold} & \textbf{start} & \textbf{end} & \textbf{occluded} \\
\midrule
rh & 4 & 0 & 208 & 0-47, 123-208 \\
lh & 4 & 260 & 358 & 260-284, 310-358 \\
rh & 4 & 260 & 382 & 260-332, 339-382 \\
lf & 0 & 292 & 465 & none \\
... & ... & ... & ... & ... \\
\bottomrule
\end{tabular}
\caption{Example of annotated frames for a single climber. The climbers' extremity used, the hold id, the duration (start and end in frames), and frame ranges with occlusions are recorded.}
\label{tab:annotatedFramesFormat}
\end{table}

\begin{figure}
    \centering
      \begin{subfigure}{0.19\linewidth}
        \includegraphics[width=\linewidth]{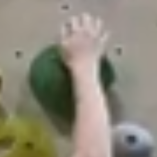}
      \end{subfigure}
      \begin{subfigure}{0.19\linewidth}
        \includegraphics[width=\linewidth]{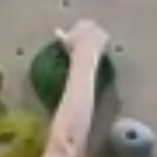}
      \end{subfigure}
      \begin{subfigure}{0.19\linewidth}
        \includegraphics[width=\linewidth]{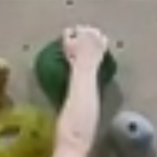}
      \end{subfigure}
      \\ \vspace{0.05cm}
      \begin{subfigure}{0.19\linewidth}
        \includegraphics[width=\linewidth]{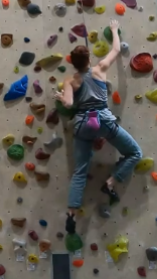}
        \caption{not touched}
        \label{fig:notTouched}
      \end{subfigure}
      \begin{subfigure}{0.19\linewidth}
        \includegraphics[width=\linewidth]{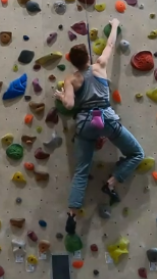}
        \caption{touched}
        \label{fig:touched}
      \end{subfigure}
      \begin{subfigure}{0.19\linewidth}
        \includegraphics[width=\linewidth]{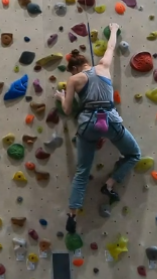}
        \caption{used}
        \label{fig:used}
      \end{subfigure}
      \caption{The climber is moving their right arm towards the hold (a), positions their hand to use the hold (b), and grabs the hold (c). Only in the case of (c) was the hold annotated as used. The top row shows closeups of the right hand.}
      \label{fig:usedVSNotUsed}
\end{figure}

The start and end frames of the hold usage denote the duration the climber held onto a hold. A hold was considered in use when the climber intentionally touched it. Incidental contact was not annotated, such as briefly brushing a hold while reaching for another. The start frame was defined as the moment the climber actively grasped or stepped on the hold. \autoref{fig:usedVSNotUsed} illustrates different stages of hand positioning on a hold, where only (c) qualifies as actual usage. The end frame was marked when the hold was released or stepped off. If the extremity was occluded at the start or end of a hold's use, body movement was analyzed to estimate when contact was made or released. 
Finally, the occlusion rate was annotated, listing frame ranges where the extremity on a hold was more than 50 \% covered by other body parts, primarily occurring for hands. 

As the camera position varies slightly across recordings, the bounding boxes of the holds in pixel coordinates are provided separately for each video. Since distinguishing dedicated footholds may be relevant, each hold also specifies this information.
To estimate real-world distances — such as the proximity of a climber’s limb to the next hold — a homography matrix is provided for each recording. While camera calibration could account for lens distortion, initial tests showed that a homography transformation was sufficient for the planar surface of the climbing wall. Homography is used in other sports to obtain game statistics on the players’ movements, including in hockey, basketball, and soccer \cite{hockeyHomography, basketballHomography, soccerfieldHomography}.
Initial tests evaluated different homography matrices using a grid of 350 bolt holes with varying numbers of reference points. The results showed that using only the four corner points yielded a mean error of 1.51 pixels — only slightly higher than the full 350-point calibration (1.20 pixels). Thus, the four-corner calibration was selected for efficiency. The dataset includes the computed homography matrix and the four reference points for each video.

\subsection*{Dataset analysis}
Our dataset has 1546.52 seconds of recorded video material from 22 videos, leading to an average of 70.30 seconds per single video. On average, the orange route was climbed faster (see \autoref{tab:video_stats}).  
\begin{table}
    \centering
    \begin{tabular}{lccc}
        \toprule
        & \textbf{green} & \textbf{orange} \\
        \midrule
        \textbf{total video duration} & 931.6 & 614.92 \\
        \textbf{avg. per video} & 84.69 & 55.90 \\
        \textbf{standard deviation} & 37.46 & 14.74 \\     
        \textbf{shortest video} & 51.2 & 38.64 \\
        \textbf{longest video} & 188.24 & 88.48 \\
        \bottomrule
    \end{tabular}
    \caption{Video duration statistics in seconds for the green and orange routes.}
    \label{tab:video_stats}
\end{table}

In total, 940 hold usages are recorded throughout all videos, where overall hand usage is higher, reflected in the number of used holds and average hold usage duration (cf. \autoref{tab:hold_stats}). Climb and hold usage duration was lower for the easier orange route.  Notably, the number of annotated hand holds is lower by two holds, compared to the foot holds, in the orange route.
Hold occlusion appeared mostly with hands and is neglectable for feet (50 vs 2 \% overall) in our data. 
\begin{table}
    \centering
    \begin{tabular}{lccc}
        \toprule
        & \textbf{green} & \textbf{orange} & \textbf{entire dataset} \\
        \midrule
        \multicolumn{4}{l}{\textbf{hold usages}} \\
        \quad total \quad \quad \quad & 496 & 444 & 940 \\
        \quad hands & 260 & 221 & 481 \\
        \quad feet & 236 & 223 & 459 \\
        \midrule
        \multicolumn{4}{l}{\textbf{avg. usage duration (s)}}\\
        \quad total & 6.16 & 4.28 & 5.27 \\
        \quad hands & 6.33 & 4.65 & 5.55 \\
        \quad feet & 5.98 & 3.90 & 4.97 \\
        \midrule
        \multicolumn{4}{l}{\textbf{avg. hold occlusion (\%)}}\\
        \quad total & 26.24 & 27.81 & 26.98 \\
        \quad hands & 46.99 & 54.91 & 50.53 \\
        \quad feet & 3.37 & 0.94 & 2.19 \\
        \bottomrule
    \end{tabular}
    \caption{Statistics of the number of hold usages, hold usage duration in seconds and occlusion percentages for green and orange route and the entire dataset.}
    \label{tab:hold_stats}
\end{table}

\section*{Hold usage detection with pose-estimation models}
\label{sec:Methodology}
We propose a time-based approach for hold usage detection using detected keypoints from out-of-the-box state-of-the-art pose-estimation models and our bounding-box-based hold annotations. 

\begin{figure}
    \centering
    \begin{subfigure}{0.2\linewidth}
        \includegraphics[width=\linewidth]{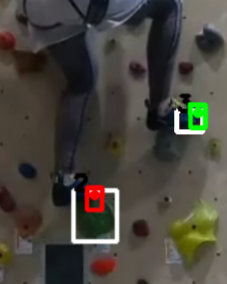} 
        \caption{ViTPose \& MediaPipe}
        \label{fig:feetVit}
    \end{subfigure}
    \begin{subfigure}{0.2\linewidth}
        \includegraphics[width=\linewidth]{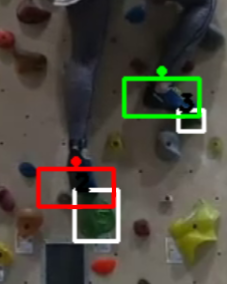}
		\caption{YOLOv8-pose}
        \label{fig:feetYolo}
    \end{subfigure}
	\caption{Different areas of interest (i.e., bounding boxes) are applied to detected foot keypoints when determining hold usage. A small area is used when the model provides the toe position (a); large areas are applied when only the ankle position is available (b). The circle represents the detected joint location, and the white rectangle indicates the annotated hold.}
	\label{fig:feetThreshold}
\end{figure}

State-of-the-art 2D pose estimation models detect hand keypoints around the wrist and foot keypoints either at the toes or the ankle. However, holds primarily engage with the fingertips and toe tips in climbing. To account for this discrepancy, an area of interest is defined around each detected keypoint to determine overlap with the marked holds, as illustrated in \autoref{fig:feetThreshold}.
The detected wrist is the center of the area of interest (i.e., an axis-aligned-bounding-box) for hand key points since hand placement on a hold can extend in any direction from the wrist. In contrast, the area of interest expands downwards for foot key points, as the foothold is predominantly below the foot. 
When the toe location is detected, a relatively small margin (see \autoref{fig:feetVit}) suffices for overlap detection. However, a significantly larger area is required for models that only estimate ankle keypoints (see \autoref{fig:feetYolo}), as the toes can be at a substantial distance from the ankle with notable movement variability.
This increased margin accounts for two factors: (1) the lateral positioning of the toes, which can shift left or right depending on foot orientation, as seen in the right foot in \autoref{fig:feetYolo}; and (2) downward foot flexion, where a pointed foot moves the toes further from the ankle, as observed in the left foot in \autoref{fig:feetYolo}. Additionally, since our tests showed that toe positions never exceeded ankle height, the entire overlap region for ankle-based models is shifted downward relative to the ankle keypoint.

The detected overlap must persist for at least 0.5 seconds (12.5 frames) for a hold to be recognized as used. This prevents brief contacts, such as when a hand or foot passes by a hold without actually using it, from being falsely classified as usage. The 0.5-second threshold was determined through empirical testing of various durations between 0 and 2 seconds, with 0.5 seconds yielding the most reliable results. A hold is considered unused when the last overlapping limb is moved.
When a climber is higher up the wall, pose estimation can be inaccurate, causing detected hand positions to become unstable. 
To account for these fluctuations, if a new overlap is detected for a hand or foot that matches a detection from a previous frame, the previous timestamp is used for the 0.5-second threshold instead of restarting from the current frame.
 
\section*{Experiments and results}
\label{sec:experimentsAndResults}

Based on a preliminary evaluation of 2D pose estimation models \cite{alphapose,openpose,YoloV8,vitpose,mediapipe}, we identified MediaPipe heavy (\textit{pose\_landmarker\_heavy}, 29.2 MB\cite{mediapipe}), YOLOv8-pose X (\textit{yolov8x-pose}, 132 MB\cite{YoloV8}), and ViTPose L (\textit{vitpose-l-coco\_25}, 1.14 GB\cite{vitpose})\footnote{For ViTPose the implementation provided by Adriano Donninelli at \href{https://github.com/JunkyByte/easy_ViTPose}{github.co/JunkyByte/easy\_ViTPose} was used.} as the most accurate models for detecting 2D poses in climbing video footage while maintaining processing speeds suitable for real-time applications. 

The implementation was done in Python (version 3.10.7) and utilizes CUDA 12.3, PyTorch (version 2.1.2+cu121), and OpenCV (version 4.8.0.76) for model application and image processing tasks. The evaluation was conducted on a laptop equipped with an NVIDIA GeForce GTX 1050 (4GB VRAM), 8GB RAM, and an Intel Core i7-8750H.
This hardware setup was selected to reflect the target use case: a portable, real-time climbing assistance tool designed to improve accessibility for visually impaired climbers. We used the downsampled video versions from the dataset to ensure feasibility on consumer-grade devices. While higher-resolution inputs may improve accuracy, lower resolutions allow for faster processing and better match real-world deployment constraints. The original 4K videos are available in the dataset \cite{ourdata}, allowing future work to explore the trade-off between resolution and performance using more powerful hardware.

To enhance the robustness of the pose estimations, the input image is cropped to the region containing the annotated holds, with an added margin to account for sideways limb movements for balance. This prevents unnecessary pose estimations of people in the peripheral areas. Additionally, previously detected keypoints are stored to mitigate potential pose estimation failures in subsequent frames.


To systematically assess model performance, we computed multiple evaluation metrics, focusing on both spatial and temporal accuracy. The evaluation process involved the following steps: (1) detecting and tracking hold usage, (2) computing accuracy, precision, and sensitivity at different granularities, and (3) analyzing the temporal alignment of hold usage detection using tIoU.

During evaluation, we recorded the detected hold-limb combinations along with their start and end frames. 
We also measured the time required to process each frame and tracked the number of individuals detected per frame. 

Accuracy, precision, and sensitivity were chosen as evaluation metrics to assess the correctness and reliability of hold detection. Sensitivity is particularly relevant, as it's sensitive to missing a used hold (false negative). Precision, on the other hand, indicates how reliable the hold detections are.
All three metrics were calculated separately for hands and feet, per route (see \autoref{tab:accuracy_per_route}) and on video level (see \autoref{tab:accuracy_participants}).
For calculating these metrics, the elements of the confusion matrix were defined as follows:
\begin{itemize}
    \item True Positive (TP): a hold that was correctly recognized as used.
    \item False Positive (FP): a hold recognized as used but not used by the climber.
    \item False Negative (FN): a hold that was not recognized as used, but was used.
    \item True Negative (TN): Not considered in this evaluation. Since most holds in a given video remain unused, including TNs, the analysis would be heavily skewed and provide little meaningful insight.
\end{itemize}
For a hold to be counted as correctly recognized (TP), the time range in which it was detected had to overlap with the time range of the ground truth usage annotation.

\begin{table*}[hb]
\centering
\begin{tabular}{|c|c|c|c|c|c|c|}
\hline
{\textbf{participant}} & \multicolumn{3}{c|}{\textbf{orange route}} & \multicolumn{3}{c|}{\textbf{green route}} \\ \cline{2-7} 
 & \textbf{ViTP.} & \textbf{Yolov8 P.} & \textbf{MP} & \textbf{ViTP.} & \textbf{Yolov8 P.} & \textbf{MP} \\ \hline
\textbf{p01} & 88.9\% & 71.9\% & \textcolor{red}{\textbf{63.3\%}} & 92.2\% & 81.9\% & 93.6\% \\ \hline
\textbf{p02a} & \textcolor{green}{\textbf{95.1\%}} & 81.3\% & 92.9\% & 95.2\% & 84.8\% & 95.2\% \\ \hline
\textbf{p02b} & 88.1\% & 72.6\% & 92.5\% & 92.7\% & 84.8\% & 88.6\% \\ \hline
\textbf{p03} & 84.6\% & 81.0\% & 82.5\% & \textcolor{green}{\textbf{97.3\%}} & 76.1\% & \textcolor{green}{\textbf{97.3\%}} \\ \hline
\textbf{p04} & 81.8\% & 60.7\% & 80.4\% & 89.4\% & \textcolor{green}{\textbf{91.3\%}} & 91.3\% \\ \hline
\textbf{p05} & \textcolor{red}{\textbf{64.7\%}} & \textcolor{red}{\textbf{50.0\%}} & 68.8\% & 83.3\% & 76.2\% & 80.5\% \\ \hline
\textbf{p06} & 74.5\% & 70.2\% & 77.1\% & 93.0\% & 81.6\% & 84.4\% \\ \hline
\textbf{p07} & 78.3\% & 72.2\% & 66.7\% & \textcolor{red}{\textbf{76.9\%}} & \textcolor{red}{\textbf{68.2\%}} & 90.4\% \\ \hline
\textbf{p08} & 83.8\% & 67.4\% & 85.3\% & 85.4\% & 86.0\% & 91.7\% \\ \hline
\textbf{p09} & 89.1\% & \textcolor{green}{\textbf{84.3\%}} & 83.7\% & 93.2\% & 88.9\% & \textcolor{red}{\textbf{69.8\%}} \\ \hline
\textbf{p10} & 93.3\% & 75.0\% & \textcolor{green}{\textbf{95.4\%}} & 93.2\% & 69.0\% & 72.3\% \\ \hline
\end{tabular}
\caption{Accuracy of different pose estimation models for each participant on the orange and green route. ViTPose L is abbreviated as ViTP., YOLOv8-pose X as Yolov8 P., and MediaPipe H as MP. The bold red values mark the worst accuracy for a model, and the bold green values mark the best accuracy for a model for a route.}
\label{tab:accuracy_participants}
\end{table*}

The tIoU is commonly used in activity recognition to determine how well the duration of an activity is detected \cite{ActivityNet}. The tIoU value for each detection is between 0 and 1, with 1 indicating perfect overlap and 0 indicating no overlap, and is calculated as follows:
\[
\text{tIoU} = \frac{|T_p \cap T_g|}{|T_p \cup T_g|}
\]
where \( T_p \) is the predicted hold usage duration and \( T_g \) is the ground truth. The tIoU was included in the evaluation as a measure of temporal alignment between predicted and actual hold usage.

Calculating accuracy, precision and sensitivity with a threshold of 0.0 for the tIoU to count a hold usage as TP, allows insights into how well hold usage detection is for cases where temporal information can be mitigated (e.g. counting used holds), while using a tIoU threshold of 0.5 provides  better insights for applications where detecting the exact time of use are of importance ( e.g. real-time next hold suggestion).
Furthermore, the time it took for one frame to be processed, the number of frames per model with no detection, and the temporal Intersection over Union (tIoU) were calculated.

\subsection*{Results}
 
\begin{table*}
    \centering
    \begin{tabular}{lccc|ccc|ccc}
        \toprule
        \textbf{tIoU\textgreater0} & \multicolumn{3}{c|}{\textbf{orange}} & \multicolumn{3}{c|}{\textbf{green}} & \multicolumn{3}{c}{\textbf{both}} \\
        & \textbf{overall} & \textbf{hands} & \textbf{feet} & \textbf{overall} & \textbf{hands} & \textbf{feet} & \textbf{overall} & \textbf{hands} & \textbf{feet} \\
        \midrule
        \textbf{ViTPose L} \\
        accuracy & \textbf{83.6\%} & \textbf{74.5\%} & \textbf{94.3\%} & \textbf{89.9\%} & \textbf{86.0\%} & \textbf{94.7\%} & \textbf{86.6\%} & \textbf{80.1\%} & \textbf{94.5\%} \\
        sensitivity & 93.7\% & 90.0\% & 97.5\% & \textbf{95.6\%} & \textbf{93.8\%} & 97.7\% & 94.6\% & 91.9\% & 97.6\% \\
        precision & \textbf{88.5\%} & \textbf{81.2\%} & \textbf{96.7\%} & \textbf{93.8\%} & \textbf{91.2\%} & 96.8\% & \textbf{91.1\%} & \textbf{86.2\%} & \textbf{96.7\%} \\
        \midrule
        \textbf{YOLOv8 pose X} \\
        accuracy & 70.9\% & 67.5\% & 74.3\% & 80.2\% & 80.7\% & 79.7\% & 75.3\% & 73.9\% & 76.8\% \\
        sensitivity & \textbf{95.4\%} & \textbf{92.0\%} & \textbf{98.8\%} & 95.2\% & 92.6\% & \textbf{98.2\%} & \textbf{95.3\%} & \textbf{92.3\%} & \textbf{98.5\%} \\
        precision & 73.4\% & 71.7\% & 75.0\% & 83.6\% & 86.2\% & 80.9\% & 78.2\% & 78.7\% & 77.7\% \\
        \midrule
        \textbf{MediaPipe H} \\
        accuracy & 80.3\% & 71.1\% & 90.9\% & 87.0\% & 82.3\% & 92.9\% & 83.5\% & 76.6\% & 91.9\% \\
        sensitivity & 92.7\% & 88.3\% & 97.1\% & 92.5\% & 89.9\% & 95.4\% & 92.6\% & 89.2\% & 96.3\% \\
        precision & 85.7\% & 78.5\% & 93.5\% & 93.6\% & 90.6\% & \textbf{97.2\%} & 89.5\% & 84.4\% & 95.2\% \\
        \bottomrule
    \end{tabular}
    \caption{Accuracy, sensitivity, and precision (tIoU\textgreater 0.0) for detecting hold usage with different pose estimation models. Separate values for the orange and green route as well as for hands and feet are provided. The best results are marked in bold font.}
    \label{tab:accuracy_per_route}
\end{table*}

ViTPose achieved the highest overall accuracy (86.6\%), followed by MediaPipe (83.5\%), while YOLOv8-pose exhibited the lowest accuracy (75.3\%). In terms of sensitivity and precision, YOLOv8-pose demonstrated the highest sensitivity (95.3\%) but the lowest precision (78.2\%). Both ViTPose and MediaPipe also exhibited higher sensitivity than precision, though the difference was less pronounced compared to YOLOv8-pose, see \autoref{tab:accuracy_per_route}.

Analyzing accuracy separately for handholds and footholds, all models achieved higher accuracy for foothold detection than for handhold detection.
Per-route accuracy analysis (\autoref{tab:accuracy_per_route}) shows that all models performed better on the green route than on the orange route. Notably, the difference in accuracy between routes was more pronounced for handholds than for footholds.
The accuracy per video varied across models. ViTPose's accuracy ranged from 64.7\% (p05) to 95.1\% (p02a) for the orange route and 76.9\% (p07) to 97.3\% (p03) for the green route, YOLOv8-pose from 50.0\% (p05) to 84.3\% (p09) for the orange route and 68.2\% (p07) to 91.3\% (p04) for the green route and MediaPipe’s from 63.3\% (p01) to 95.4\% (p10) for the orange route and 69.8\% (p09) to 97.3\% (p03) for the green route.


\begin{table}
    \centering
    \begin{tabular}{lccc}
        \toprule
        \textbf{tIoU\textgreater0.5} & \textbf{orange} & \textbf{green} & \textbf{both} \\
        \midrule
        \textbf{ViTPose L} \\
        accuracy & \textbf{68.0\%} & 70.6\% & \textbf{69.3\%} \\
        sensitivity & 76.4\% & 75.1\% & 75.7\% \\
        precision & \textbf{86.0\%} & 92.2\% & \textbf{89.1\%} \\
        \midrule
        \textbf{YOLOv8 pose X} \\
        accuracy & 59.4\% & 66.3\% & 62.8\% \\
        sensitivity & \textbf{80.6\%} & \textbf{78.9\%} & \textbf{79.7\%} \\
        precision & 69.3\% & 80.6\% & 74.7\% \\
        \midrule
        \textbf{MediaPipe H} \\
        accuracy & 64.6\% & \textbf{71.8\%} & 68.2\% \\
        sensitivity & 75.1\% & 76.4\% & 75.7\% \\
        precision & 82.3\% & \textbf{92.4\%} & 87.3\% \\
        \bottomrule
    \end{tabular}
    \caption{Accuracy, sensitivity, and precision (tIoU\textgreater0.5) of different pose estimation models on the orange route, green route, and overall. The best results are marked in bold font.}
    \label{tab:accuracy_per_route_05thrshold}
\end{table}

The mean tIoU was 0.70 for ViTPose, 0.71 for MediaPipe, and 0.73 for YOLOv8-pose. The accuracy, sensitivity and precision when using a 0.5 threshold for the tIoU to classify a hold usage as successful detection, see \autoref{tab:accuracy_per_route_05thrshold}, are noticeably worse than those for a threshold of 0.0 for tIoU (\autoref{tab:accuracy_per_route}).
Regarding inference speed, MediaPipe was the fastest, processing frames at an average of 0.12 seconds per frame (8.19 FPS). YOLOv8-pose processed frames at 0.15 seconds per frame (7.06 FPS), while ViTPose was the slowest at 0.25 seconds per frame (4.11 FPS), taking nearly twice as long as MediaPipe.
Some frames lacked any detected poses. MediaPipe had the highest number of missing detections, with over 150 frames affected in three videos. YOLOv8-pose had over 35 missing detections in two videos and between 1 and 15 frames in eleven videos. ViTPose missed between 2 and 10 frames in seven videos.

Overall ViTPose L, and MediaPipe H, show accuracies above 80\%. YOLOv8-pose X delivers the worst accuracy with 75.3\%. Sensitivity is higher than precision for all models, suggesting that they rarely miss actual hold usage (low false negatives) but tend to overpredict usage (higher false positives). This pattern is most pronounced in YOLOv8-pose X, likely due to its reliance on ankle keypoints, requiring bigger bounding boxes for foot overlap checks, which increases the likelihood of misclassifications.
Looking at the accuracies for the foot hold usage detection we can notice that the accuracy for the feet for this model with 76.8\% is significantly worse than the accuracies for the feet for both other models, which are above 90\%. This suggests that the keypoint choice, and the size of the area of interest, greatly impact accuracy. 
Notably, the accuracy for the handhold usage detection was worse for YOLOv8-pose X as well with 73.8\%, however not as big with 76.6\% (MediaPipe H) and 80.1\% (ViTPose L).
Comparing the accuracy of foothold usage detection with handhold usage detection for all models, we notice, that foothold usage accuracies are better for all three models. Longer occlusions of the hands behind the climber’s upper body frequently lead to detection inaccuracies, whereas brief leg occlusions (e.g., one leg covering the other) appear to have minimal impact on accuracy.

Looking at the accuracies separated by route in \autoref{tab:accuracy_per_route}, we can see that all accuracies (overall, hand and feet) for the green route are better for all models. Interestingly, the accuracies for the handhold usage detection improved a lot more for all models than the accuracies for the feet. The green route's holds are further apart, requiring climbers to use dynamic movements. This likely reduces the frequency of hand occlusions by the upper body, improving wrist keypoint detection and leading to higher handhold accuracy.

We hypothesized that experienced climbers, who climb with more extended arms, would show higher pose estimation accuracy than beginners, who tend to pull close to the wall. However, accuracy did not correlate with experience, suggesting that factors like the specific route, camera angle or pose estimation challenges at higher wall sections are more influential.

Interestingly, MediaPipe H had issues detecting poses in more than 100 consecutive frames in three cases, resulting in low accuracy at video level, for all three cases (63.3\%, 69.8\%, 72.3\%). When looking at the frames where the model started failing to predict the keypoints, in all three cases the climber was higher up the wall and the head of the climber was hidden behind their upper body, as they were looking down at their feet.

Looking at the mean tIoU, YOLOv8-pose shows the highest values followed by MediaPipe and ViTPose with the lowest values. Accurately predicting the exact time intervals of hold usage remains challenging. The low values may be due to the number of frames processed, reducing the models ability to precisely capture hold usage timing. With fewer frame updates, the model has fewer opportunities to capture the exact moment a hold is used or released, leading to minor misalignments in hold duration predictions.

This results in trade-offs in model selection:
\begin{itemize}
    \item \textbf{MediaPipe H} is the fastest model with high accuracies, however shows occasional pose detection failures when the head of the climber is covered by their upper body.
    \item \textbf{ViTPose L} is the most accurate model with rare failure in detecting a pose, however processing times are twice as long.
    \item \textbf{YOLOv8-pose X} has the worst accuracy results, largely due to the use of ankle keypoints instead of toe keypoints.
\end{itemize}

\section*{Conclusion}
\label{sec:conclusion}

In this work, we introduced a novel dataset for hold-usage detection in sports climbing and propose using state-of-the-art pose estimation models to detect hold usage.   
We assessed the application of three 2D pose estimation models, MediaPipe, ViTPose, and YOLOv8-pose, on our dataset and identified key challenges, such as self-occlusions of body parts and skeleton conventions impacting detection performance. 
Overall, the best performing model, ViTPose, achieved an accuracy of 86 \% while processing frames at interactive rates. If high frame rates are a requirement, the second-best model in terms of accuracy, MediaPipe, still achieves 83 \% while doubling the frame rate compared to ViTPose. 
Our findings highlight the potential of state-of-the-art pose estimation models for applications in the emerging sport of climbing.

\subsection*{Future work}
In the future, we plan to investigate the impact of other skeleton conventions on hold usage detection. Models with key points at the fingertips can potentially improve our results significantly.

While our dataset \cite{ourdata} provides a valuable foundation by focusing on planar sport climbing walls with consistent verticality, future extensions could incorporate additional recordings featuring varied wall angles and heights. This would enhance the dataset's diversity and support broader generalization across different climbing scenarios.

We aim to integrate our findings into a vision-based system to assist visually impaired climbers by detecting hold usage and predicting subsequent holds. Currently, visually impaired climbers face challenges, either navigating unspecified routes with limited training efficacy or relying on sighted guides, which restricts training flexibility. An automated system that detects the climber's position, predicts the next hold, and provides auditory guidance could significantly improve training.

As an initial step, we explored predicting a climber's upward reach to potential holds based on current hand positions and pose. While promising, this approach is not yet sufficient for real-world applications, highlighting the need for further research into a more reliable next-hold recommendation system.






\bibliography{main}

\end{document}